\documentclass[runningheads]{llncs}

\usepackage[T1]{fontenc}
\usepackage{graphicx}
\usepackage{algorithm}
\usepackage{algpseudocode}
\usepackage{svg}
\usepackage{array}
\usepackage{hyperref}
\usepackage{todonotes}
\begin{document}

% Titulos:
% Natural Language Optimizer for Retrieval-Augmented Generation
% Optimizing Retrieval Augmented Generation with Meta-Prompting
% Retrieval Augmented Generation Prompt Optimizer with Large Language Models
% Automatic Refining Prompts for Retrieval Augmented Generation through Iterative Meta-prompt Optimization
% Optimization of Prompts for Enhanced Retrieval-Augmented Generation
% Content Optimization Meta-Prompting: Advancing Retrieval-Augmented Text Generation
% Enhancing Retrieval Augmented Generation with Meta-Prompting
% Post-Retrieval Optimization with Meta-Prompting
\title{Meta-prompting Optimized Retrieval-augmented Generation}

\author{João Rodrigues \and
António Branco}
\institute{University of Lisbon\\
NLX---Natural Language and Speech Group, Dept of Informatics\\
Faculdade de Ciências (FCUL), Campo Grande, 1749-016 Lisboa, Portugal\\
\email{\{jarodrigues,antonio.branco\}@fc.ul.pt}}

\maketitle            

\begin{abstract}
Retrieval-augmented generation resorts to content retrieved from external sources in order to leverage the performance of large language models in downstream tasks.
%, dispensing with re-training it for up-to-date knowledge to be taken into account.
%, furthermore, it can pinpoint the sources from which the generated responses are based improving thus its credibility. 
%, another advantage is the possibility of using proprietary data and consequently diminishing security problems.
The excessive volume of retrieved content, the possible dispersion of its parts, or their out of focus range may happen nevertheless to eventually have a detrimental rather than an incremental effect.
%typical retrieval-augmented generation pipeline feeds a large language model with the concatenation of the retrieved external sources content without a connecting thread which may hold back performance. 
%For instance, rewriting the retrieved external source's contents may improve retrieval-augmented generation.
To mitigate this issue and improve retrieval-augmented generation, we propose a method to refine the retrieved content before it is included in the prompt by resorting to meta-prompting optimization.
Put to empirical test with the demanding multi-hop question answering task from the StrategyQA dataset, the evaluation results indicate that this method outperforms a similar retrieval-augmented system but without this method by over 30~\%.

\end{abstract}

\section{Introduction}

Pre-trained Large Language Models (LLMs) \cite{vaswani2017attention,Raffel:2020:T5} are known for their hallucinations \cite{ji2023survey} and for their further limitations regarding truthfulness \cite{lin2022truthfulqa}. To tackle these issues, remediation techniques have been explored such as, for instance, fine-tuning \cite{howard2018universal}, prompt-engineering \cite{liu2023pre} or Retrieval-Augmented Generation \cite{lewis2020retrieval}, initiated by Houlsby et~al. \cite{houlsby2019parameter} among several others.

\subsection{Retrieval-augmented generation}

% https://app.diagrams.net/
%\begin{figure}[tp]
%\centering
%\includegraphics[scale=0.4]{images/rag.jpg}
%\caption{\textbf{Retrieval-augmented generation (RAG)} - A typical RAG pipeline where users submit a query and obtain a response.}
%\label{fig:rag}
%\end{figure}

Focusing on Retrieval-Augmented Generation (RAG), this approach seeks to enhance truthfulness and curb hallucinations by expanding the initial prompt, which contains the initial query, with additional content retrieved from sources that are external to the LLM. Such additional content is obtained with the help of an auxiliary Retrieval Model 
%--- see Figure \ref{fig:rag} for a RAG scheme --- 
where the retrieval model may be a simple Jacquard model or a vector database that extracts relevant content from external sources and pass it on to a Large Language Model that generates an appropriate response given the original query and the extracted content.
If this external content is unstructured text, it may be of different lengths, such as sentences, paragraphs or full documents, among others. 

%As an example to briefly illustrate RAG, consider the following initial query:

%\begin{itemize}
%    \item Would nickel boil if placed in the core of Earth?
%\end{itemize}

%It could be used to retrieve relevant content from external sources, for instance like the following example sentences: 

%\begin{itemize}
%    \item The boiling point of nickel is 3003 Kelvin.
%    \item The temperature of earth's outer core is 3,000–4,500 Kelvin.
%\end{itemize} 

%Next, both the initial query and the retrieved content are included in a prompt, the retrieval-augmented prompt, which is entered into an LLM for this to generate a response. Given this enhanced prompt that contains further context with potentially more sharp information, very likely the response obtained is better than what would have been obtained with just the initial query.
%Continuing with our example, hopefully, the generated response could be:

%\begin{itemize}
%\item Yes, nickel boil if placed in the core of Earth.
%\end{itemize}

By feeding LLM's knowledge, and curbing its whim, with further knowledge from external sources, more accurate answers are likely to be provided.

Compared with other techniques, such as fine-tuning or prompt-engineering, RAG key advantage is the ease with which newer, up-to-date content is taken advantage of, as this does not require the costly compute of re-training neural networks (as in fine-tuning) or the costly human labour for the creation of further manually designed prompts (as in prompt-engineering).
To be sure, all these techniques can nevertheless be mixed and function together.

\subsection{Prompt optimization}

Usually, the pieces of content retrieved may be from heterogeneous sources and they tend to lack a connecting thread.
They may also be redundant or may be of very high volume.
These, among other aspects, may end up having a detrimental effect and eventually jeopardizing the generation task, rather than enhancing it.

To mitigate this problem, we present a method that consists of adding an intermediate step between the retrieval of the external content and the entering of the expanded prompt into the LLM to finally obtain the response to the initial query. Aiming at improving the performance of this generation-LLM, this intermediate step seeks to obtain a refined version of the external knowledge.

This refinement is accomplished by means of an auxiliary transformation-LLM that is entered with a prompt containing the pieces of retrieved contents, preceded by an instruction with the request for the sought refinement. 

For example, if several pages of Wikipedia are retrieved as possible relevant content, the transformation-LLM processes this content and may generate a summary or remove unnecessary information from that original content.

%That is, our method applies a prompt optimization method.
%This is a solution to the laborious task of manually crafting prompts, which can be expensive and may require repetition whenever the external data sources are updated.

%To automatically search for a prompt that effectively transforms the retrieved content, we applied an LLM as an optimizer.
%We begin by defining in natural language the optimization task, referred to as the meta-prompt.

Turning to this refinement instruction, this is obtained by an automatic procedure that is preliminary to running the RAG system made of transformation- and generation-LLMs, and it is undertaken by yet a third LLM.

Inspired in Yang et~al. \cite{yang2023large}, in this procedure a meta-prompt is used as input to this third, optimizer-LLM for this to iteratively generate new tentative instructions, score them, and retain, in the meta-prompt itself, a list with the top k ones that induce better performance for the RAG system. By the end of this optimization process, the best scoring instruction in this list is the one retained to be used in the refinement step of the retrieved contents with the transformation-LLM.

This meta-prompt contains a meta-instruction and a list of tentative instructions that is aimed at being updated during this process with new instructions that induce better RAG performance.
After a new tentative instruction is generated, its contribution to approximate the gold output to the initial query is scored, and the list of tentative instructions in the meta-prompt is possibly updated so that it retains the top-performing ones so far.
This is iterated, and an optimization trajectory is hence accomplished to eventually find the new refinement instruction that maximizes the success of the RAG system.

%The transformation in (ii), in turn, is accomplished by passing on the initially retrieved content to an LLM (transformation-LLM), which returns the retrieved content transformed. 

%Given that LLMs are highly sensitive to various prompts (cite), and the optimal prompt is both model-specific and task-specific, (cite) determining an effective prompt for the transformation-LLM is pivotal.

\begin{center}***
\end{center}

In this paper, we propose a method for RAG to be enhanced with the refinement of the retrieved content, a refinement that is optimized by resorting to iterative meta-prompting. This is a novel method that can be combined with previous approaches aimed at enhancing RAG.

We report on the experiments performed to put this method to the test. This approach is extrinsically evaluated by being embedded in a demanding question-answering downstream task. Its performance demonstrates that it is an effective method to enhance RAG by improving by 30\% the performance of a baseline RAG without this method, and that it can be combined with other previous state of the art methods for RAG enhancement proposed in the literature.

The remainder of this paper is structured as follows:
Section~\ref{sec:rel_work} discusses related work; 
Section~\ref{sec:method} describes the method proposed in this study;
Section~\ref{sec:models} reports on the models and dataset resorted to;
Section~\ref{sec:discussion}
presents the
experiments undertaken and their evaluation, and discusses the results obtained;
Section~\ref{sec:future} addresses future research paths;
and finally, Section~\ref{sec:conclusion} closes this paper with concluding remarks.

%We would like to emphasize that our method operates within the framework of optimization constrained to the use of natural language for describing the optimization task.  However, it's important to note that there may exist more efficient formal methods beyond the scope of our approach.

\section{Related work}
\label{sec:rel_work}

Prompt optimization has gained traction as an effective mechanism for enhancing LLMs in several downstream tasks \cite{srivastava2023beyond}\cite{lester2021power}.

The earliest approaches in prompt optimization sought to directly optimize the prompt embedding space, such as prefix-tuning \cite{li2021prefix} or OptiPrompt \cite{zhong2021factual}. These aimed at optimizing a sequence of continuous task-specific vectors applied to the prompt to leverage downstream tasks.

More recent studies have introduced further techniques to enhance prompts, such as chain-of-thought \cite{wei2022chain} and tree-of-thoughts \cite{yao2024tree}.
The former involves extending prompts with a few manually written chain of thought demonstrations as examples, which results in improved performance across various tasks, including arithmetic, commonsense and symbolic reasoning. 
The latter builds upon the chain-of-thought by considering multiple reasoning paths, self-evaluating choices, and by making global decisions by looking ahead or backtracking when necessary.

Other methods for optimizing prompts include searching through a pool of prompt candidates generated by an LLM, %\cite{zhou2022large}, 
employing principled planning algorithms based on Monte Carlo tree search \cite{wang2023promptagent}, or applying iterative local edit operations at a syntactic phrase-level split within the prompts \cite{prasad2023grips}. 

Further proposals encompass EvoPrompt \cite{guo2023connecting}, which uses evolutionary operators over a prompt population for optimization,  while Sabbatella et~al.  employs Bayesian Optimization within a prompt search space \cite{math12060929}, reinforcement learning to rewrite prompts \cite{kong2024prewrite} or a prompt optimization that integrates human-design feedback rules to suggest improvements automatically \cite{chen2024prompt}.

Recently, Yang et~al. \cite{yang2023large} introduced OPRO, leveraging LLMs as optimizers through meta-prompts, which are natural language descriptions that guide prompt optimization.
It was applied to optimize prompts by retrieving and re-ranking top-K relevant instructions with respect to an initial instruction, and by appending them to the global task description.

In contrast, to enhance RAG, we propose a method to optimize the prompt that differs from the previous proposals in the literature.

A prompt for RAG includes a query and the content retrieved from external sources on the basis of that query. It may contain also an instruction about how to handle the query or how the retrieved content should be used by the generation-LLM to answer it. Related work for RAG enhanced with prompt optimization has concentrated on optimizing the instruction and/or the query. Differently from previous approaches, our method focuses instead on optimizing the version of the retrieved content that is included in the prompt entered into the generation-LLM. Hence, rather than being an approach alternative to previous ones, it is a new one that is complementary to them and may be combined.

%our proposal extends beyond these previous approaches by focusing on searching for a prompt that rewrites the retrieved contents.

\section{Method}
\label{sec:method}

%The objective is to obtain an optimal prompt that will be entered to an LLM and rewrites the retrieved content aiming to improve a RAG task. 
%To provide a clear description of our approach it follows a description of the RAG task used to evaluate the downstream performance of the method, the description of our method. 

The objective of our method is to enhance the RAG performance of a generation-LLM by means of the improvement of its input prompt, which is made of a query introduced by the user and of pieces of content retrieved from external sources on the basis of that query. Before it is entered into the generation-LLM, this prompt is improved by means of a refinement of the retrieved content, performed by a transformation-LLM.

%\subsection{Method}

\begin{table}[]
\scriptsize
\begin{tabular}{|p{0.5cm}p{11cm}p{0.5cm}|}
\hline

& \multicolumn{1}{c}{} &  \\
& 
I have some prompts along with their corresponding scores. The prompts are arranged in ascending order based on their scores, where higher scores indicate better quality. Together with relevant information extracted from a database, these prompts are given as input to a large language model in order to optimize the provided relevant information. Several techniques may help the optimization, such as re-ranking paragraphs, cleaning, filtering and summarization. Write your new prompt taking into account the previous ones and aiming to achieve a higher score. 
\newline
\newline
{\color[HTML]{009901} 
%prompt:\newline
%Clean and organize the previous text.\newline
%score:\newline
%0.5\newline
%\newline
%prompt:\newline
%Rewrite the previous text with a more academic tone.\newline
%score:\newline
%1\newline
%\newline
prompt:\newline
Summarize the main idea of the previous text.\newline
score:\newline
3.0\newline
\newline
prompt:\newline
Summarize the main points in 30 words or less.\newline
score:\newline
3.0\newline
}

                      & \\
 \hline
\end{tabular}
\newline
\caption{\textbf{Meta-prompt} - An example of a meta-prompt: in black, the top paragraph with the meta-instruction actually used in the experiments; below, in green, the list of top performing instructions so far, and the respective scores.}
\label{tab:meta_prompt_example}
\end{table}

And before a first query is accepted to put the RAG system to use, the prompt to be used with the transformation-LLM for refinement purposes is optimized. This prompt includes a refinement instruction and the pieces of retrieved content to be refined. It is optimized by means of the optimization of this instruction through iterative meta-prompting.

This meta-prompting optimization is undertaken by an optimizer-LLM that is entered with a (meta-)prompt that includes a (meta-)instruction and a list of tentative refinement instructions and respective performance scores. These scores are obtained by running the RAG with the tentative refinement instruction through a sample of training examples and evaluating the output against the respective gold responses.

Focusing on the optimization phase, a meta-prompt is used that contains both the description of the optimization problem and the history with previous best solutions for the instruction. Such meta-prompt is iteratively entered into the optimizer-LLM, and at each iteration that history is possibly updated with generated instructions if these support better performance for the task at stake in the generation phase. The instruction selected out of this optimization process is the best scoring one in the history obtained as this iteration is over.

An example of a meta-prompt is in Table \ref{tab:meta_prompt_example}, and a detailed description of this optimization via iterative meta-prompting is presented in Algorithm \ref{alg:meta}.

\vspace{-0.5cm}
\begin{algorithm}[]
\caption{Optimization with meta-prompting}\label{alg:meta}
\begin{algorithmic}[1]
\State \textbf{Input:} Dataset $D$ with $n$ examples, each containing a query $q$, retrieved contents $c$ and the answer $a$; meta-prompt $metaP$ with the description of the optimization task and with a list of instructions and respective scores
\State \textbf{Output:} List of scored instructions and the best scoring instruction

\While{optimizing prompt}
    \State Enter meta-prompt to optimizer-LLM 
    \State Generate new instructions $I$
    \State Select a random subset $E$ of examples $e$ from $D$
    \For{each instruction $I_j$}
        \For{each example $e_k$ in $E$}
            \State Assemble prompt $TransP$ from $I_j$ and contents $c_k$ 
            \State Enter $TransP$ to transformation-LLM
            \State Generate transformed contents $tc$
            \State Assemble prompt $TaskP$ from query $q_k$ and $tc$
            \State Input $TaskP$ to generation-LLM
            \State Generate answer and evaluate it against gold $a_k$
        \EndFor
        \State Compute $I_jscore$
    \EndFor
    \State Update $metaP$ by replacing its worst scoring instruction by $I_j$ and $I_jscore$ if this is better scored
\EndWhile
\end{algorithmic}
\end{algorithm}
% \vspace{-1cm}

\section{Dataset and models}
\label{sec:models}

To empirically assess the performance gains of the proposed method, it was integrated into an RAG for question-answering whose performance provides for its extrinsic evaluation.

\subsection{Task and dataset}
% strategyQa
% https://direct.mit.edu/tacl/article/doi/10.1162/tacl_a_00370/100680/Did-Aristotle-Use-a-Laptop-A-Question-Answering

%We used a RAG question-answering task to evaluate our method.

Multi-hop question answering requires taking into account disparate pieces of content to get at the answer for a query, which constitutes a most demanding scenario for the task of question answering.

We resorted to a most complex benchmark for multi-hop question-answering available in the literature, the StrategyQA dataset \cite{geva2021did,min-etal-2019-compositional,inoue-etal-2020-r4c,ho-etal-2020-constructing}, which contains 2,780 queries, each associated with related content made of paragraphs and the respective \textit{yes} or \textit{no} answer. 
Based on Wikipedia content, this dataset covers a range of diverse topics and
the task consists in, given a query, to provide an accurate answer to it together with the passages retrieved from Wikipedia with the most relevant content to get at that answer --- Table \ref{tab:strategyqa_example} displays an example. %, where the most relevant contents are already selected.

%StrategyQA contains 2,780 queries, where each is associated with related content made of paragraphs and a \textit{yes} or \textit{no} answer.
%To answer a query the most relevant content must be retrieved and reasoned over.

\begin{table}[h]
\scriptsize

    \centering
    \begin{tabular}{p{3cm}p{9cm}}
    \hline
    
    \textbf{Query}   & Could \$1 for each 2009 eclipse buy a copy of TIME magazine in 2020?                    \\ \hline
    \textbf{Content \#1} & It set out to tell the news through people, and for many decades through the late 1960s, the magazine's cover depicted a single person. [...]
%    More recently, Time has incorporated "People of the Year" issues which grew in popularity over the years. Notable mentions of them were Barack Obama, Steve Jobs, etc. The first issue of Time was published on March 3, 1923, featuring Joseph G. Cannon, the retired Speaker of the House of Representatives, on its cover; a facsimile reprint of Issue No. 1, including all of the articles and advertisements contained in the original, was included with copies of the February 28, 1938 issue as a commemoration of the magazine's 15th anniversary. The cover price was 15\textcent (equivalent to \$2.25 in 2019). On Hadden's death in 1929, Luce became the dominant man at Time and a major figure in the history of 20th-century media. According to Time Inc.: The Intimate History of a Publishing Enterprise 1972-2004 by Robert Elson, "Roy Edward Larsen [...] was to play a role second only to Luce's in the development of Time Inc". In his book, The March of Time, 1935-1951, 
Raymond Fielding also noted that Larsen was "originally circulation manager and then general manager of Time, later publisher of Life, for many years president of Time Inc., and in the long history of the corporation the most influential and important figure after Luce"                                                        \\ \hline
    \textbf{Content \#2} & Total eclipses are rare because the timing of the new moon within the eclipse season needs to be more exact for an alignment between the observer (on Earth) and the centers of the Sun and Moon. [...]
%    In addition, the elliptical orbit of the Moon often takes it far enough away from Earth that its apparent size is not large enough to block the Sun entirely. Total solar eclipses are rare at any particular location 
because totality exists only along a narrow path on the Earth's surface traced by the Moon's full shadow or umbra.                                                                               \\ \hline
    \textbf{Content \#3} & At least two lunar eclipses and as many as five occur every year, although total lunar eclipses are significantly less common. If the date and time of an eclipse is known, the occurrences of upcoming eclipses are predictable using an eclipse cycle, like the saros.
    \\ \hline
    \textbf{Answer}     & Yes 
    \\ \hline
    \end{tabular}
    \newline
    \caption{\textbf{StrategyQA} - An example from the StrategyQA, with a query, three of the most relevant pieces of content, and the respective answer. }
    \vspace{-1cm}
    \label{tab:strategyqa_example}
\end{table}

To provide for the evaluation of the proposed method, and isolate the accrued performance induced by it, thus disregarding possible fluctuation or loss of performance due to the retrieval process, only the gold pieces of content from a test set should be taken into account.
Since the answers are not provided in the original test set of StrategyQA, a new test set for the present evaluation exercise had to be built.
Accordingly, we divided the original training set into two parts: a new test set with 490 of the original training examples, which matches the size of the original test set, and a new training set containing a subset with 1800 such examples. The resulting train and test sets have an average query length of 9.6 words, and 2.33 contents (paragraphs) per query and are almost balanced.
The training set contains 834 yes answers and 966 no answers (46.32~\% / 53.68~\%). 
The test set contains 237 yes answers and 253 no answers (48.40~\% / 51.60~\%).

\subsection{Models}

Two Transformer-based language models with 70 Billion parameters were used, a pre-trained Llama-2-70b and an instruct model Llama-2-70b-chat fine-tuned for dialogue use \cite{touvron2023llama}.
These models were trained and fine-tuned with a context length of 4k tokens over 2 Trillion tokens on a mix of publicly available data.

% Given the constraints of hardware, the hyper-parameters regarding the generation context length were limited.

% hyperparameters

In general, the default Llama2 model hyper-parameters were applied and no hyper-parameters search bound was performed.
All language models use a temperature value of 1.0, a maximum of 64 generation tokens for the new instructions, a maximum of 128 generation tokens for the refined content, and a maximum of 64 generation tokens for the response to the task.
The optimization run was performed for two days on two NVIDIA A100 40GB GPUs.

All software and versioning along with hyper-parameters are fully described in the source code of these experiments.\footnote{For the sake of reproducibility, data and code are available at \url{https://github.com/nlx-group/rag-meta-prompt}}

\subsection{Evaluation procedure and metrics}

Based on empirical experimentation, we arrived at a meta-prompt, presented in Table \ref{tab:meta_prompt_example}, that indicates the aim of the optimization problem and includes a starting example instruction.\footnote{The starting instruction is "Clean and organize the previous text."}

The instruction optimization was iterated over 100 steps. At each step, 3 instructions were generated, each such instruction was evaluated on a random sample of 6 training examples, and the meta-prompt was eventually updated to containing the 8 top scoring queries so far. When this iteration was concluded, the best scoring instruction was retained as the optimized instruction.

We compare against the same generation-LLM using test queries and associated pieces of content, that is without the later being refined by the transformation-LLM under the instruction that was optimized by the optimizer-LLM.%, thus, skipping the optimization of prompts and the retrieved content transformation.
% Comparison with human prompts for this task is not available in the literature. SERÁ? ver related work

%To evaluated the proposed method, and focusing on its possible added value, only the gold pieces of content from the test set were taken into account for these experiments.

For the StrategyQA task, a Boolean answer is expected. Accuracy is thus the metric used for evaluating the match between the answer output by the system and the gold answer in the data set.
%In the evaluation of the question answering system, the query coming out of the query optimization process with the highest score is used.
Accuracy score is given by the proportion of correct answers, and a generated response was counted as correct if the gold answer was found in the exact beginning of it.
The response underwent minimal normalization, with just lowercasing.

 %For each example in the test set, the retrieved contents undergo transformation, followed by evaluation with the corresponding example query in the question-answering task.

As for the instrumental process of instruction optimization, it is worth recalling that the evaluation is performed over sample examples from the training set. For a tentatively generated instruction, a correct answer to it counted 1 point; an incorrect answer, in turn, counted 0.5 points if it was nevertheless in a Boolean format, or counted 0 points otherwise.
The maximum possible score was thus 6 points, given each tentative instruction was evaluated against 6 sampled queries as indicated above.

%During the optimization process, the evaluation function evaluated prompts using an exact match over the response, if correct add 1 point, if incorrect only add 0.5 points if the answer is in a boolean format. % explicar pq 0.5 points? queriamos dar pontuação a prompts que têm em conta o tipo de avaliação

\section{Results and discussion}
\label{sec:discussion}

In this section, we report on the evaluation exercise undertaken to assess the proposed method and discuss its results, summarized in Table \ref{tab:results_strategyqa}

\subsection{Experiments}

All in all, six experiments were undertaken, two resorting to the model Llama-2-70b, developed with a pre-training regime only, and four resorting to the model Llama-2-70b-chat, which resulted from further fine-tuning it with dialogue data.

\begin{table}[h]
\centering
\begin{tabular}{wl{3cm}wl{5cm}l}
\hline
\multicolumn{1}{c}{\textbf{Model}} & \multicolumn{1}{c}{\textbf{Method}} & \textbf{Accuracy} \\ \hline
Llama-2-70b      & query                        & 17 (3.46 \%)   \\
Llama-2-70b      & query+contents               & 33 (6.73 \%)   \\
\hline
Llama-2-70b-chat & query                        & 81 (16.53 \%)  \\
Llama-2-70b-chat & query+contents (plain RAG)               & 128 (26.12 \%)  \\
Llama-2-70b-chat & refined query+contents (ours) & \textbf{170 (34.69 \%)} \\
Llama-2-70b-chat & ref. query+contents no iteration       & 127 (25.92 \%)  \\
\hline
\end{tabular}
\newline
\caption{\textbf{Evaluation} - From the total 490 test set examples, the number of correct answers is presented and the respective accuracy.}
\vspace{-0.8cm}
\label{tab:results_strategyqa}
\end{table}

Both these models were used in two evaluation scenarios. In one of these scenarios---noted as \textbf{query} in Table \ref{tab:results_strategyqa}---, the response to the query entered was provided by the LLM alone, with no further content from external sources being entered. In the other scenario, in turn,---noted as \textbf{query+contents}---, further content from external sources was included in the prompt as well. The performance scores for these two scenarios with the two models are displayed in the top four rows \ref{tab:results_strategyqa}.

\textbf{External, non-parametric content improves generation} As expected, and in line with results in the literature, the retrieval-augmented generation (26.12\%) outperforms the plain generation based solely on the query (16.53\%).

\textbf{Fine-tuning improves generation} Also as expected, and by a very large margin, better performance scores were obtained with Llama-2-70b-chat, which had been fine-tuned on dialogue tasks, namely 16.53\% against 3.46$\%$, with the query only, and 26.12\% against 6.73\%, with the query and external content.

\textbf{Retrieved content refinement via meta-prompting optimization improves RAG --- the proposed method is effective}
The model Llama-2-70b-chat was thus retained and two further evaluation scenarios were considered.

A scenario with the application of the proposed method---noted as \textbf{refined query+contents}---, where the external content was refined with the help of an instruction optimized with meta-prompting.

The performance scores indicate that, with 34.69\% accuracy, our proposed method of enhanced RAG outperforms plain RAG, with 26.12\%, thus contributing for a large improvement of over 8.5 percentage points, that represents here an improvement rate of almost 33\%.

\textbf{Retrieved content refinement via "brute force" optimization does not improve RAG}
A sixth scenario---noted as \textbf{refined query+contents no iter}---was also considered. Here the external content was refined as in the proposed method, but the instruction was refined under an alternative way that dispensed with iterative meta-prompting.

All in all, 300 tentative instructions are generated during all optimization steps ---recall that we had 100 iteration steps with 3 tentative instructions generated per step with meta-prompting optimization. To dispense with this iterative meta-prompting, the same number of 300 new tentative instructions were generated at once, in a "brute force" fashion.
By the end of this process, all tentative instructions were scored with the same scoring function as in the proposed method, and the top instruction was evaluated on the test set.

This "brute force" optimization approach, scoring 25.92\%, is outperformed not only by the proposed method of meta-prompting optimization, with 34.69\%, but even also by the baseline, plain RAG, with 26.12\%.

%Legend for the methods: \textbf{query} - the query of each test set example; \textbf{contents} - the gold retrieved contents regarding the query; \textbf{transformed contents} - the contents resulting from the application of the best-found transformation prompt; \textbf{brute force generation} - includes the query and the contents resulting from the application of the best-found transformation prompt through the brute force generation of prompts without optimization.

%The results for the meta-prompt optimization as well as the results of the subsequent analysis (described in Section \ref{sec:discussion}) results are presented in Table \ref{tab:results_strategyqa}. 

\textbf{Statistical significance}
To assess the statistical significance of the improvements by our method, we employed the unpaired t-test.\footnote{The unpaired t-test evaluates if there exists a statistically significant distinction between the means of two independent samples by comparing them.}
% baseline 
% 128/490 (26.122448979591837)
% 122/490 (24.897959183673468)
% 119/490 (24.285714285714285)
% optimization 
% 170/490 (34.69387755102041
% 183/490 (37.3469387755102)
% 172/490 (35.10204081632653)
We evaluated the baseline system with three seeds and did the same for the meta-prompting optimized system.
Both samples are independent and one may assume the samples are normally distributed.
% https://www.graphpad.com/quickcalcs/ttest2/
Applying the unpaired t-test, a two-tailed P value equal to 0.0004 was obtained, which is considered statistically significant.

\subsection{Examining the tentative instructions}

%We improved a retrieval-augmented generation question-answering task by optimizing a prompt to enhance retrieved content from external sources.
%These results are indicated in Table \ref{tab:results_strategyqa} when comparing the accuracy obtained without the meta-prompt optimization (26.12 \%), fifth row, against the results obtained by our method (34.69 \%), sixth row. 
%\textbf{An improvement of 8.57 \% accuracy is obtained with our method.}

Table \ref{tab:top_prompts} presents the top generated prompts. The best scoring prompt (last row), with 5.5 (out of a maximum of 6), was obtained at iteration step 46 (out of 300 steps in total), and a good prompt (first row) can be obtained with only 28 steps.

When reading the best prompt (last row), one realizes that it aims to improve the task through the summarization of the retrieved contents, considering their broader context, and identifying the main theme or message. 
It appears thus like a reasonable prompt a human might have thoughtfully arrived at if aiming at improving the performance of the task.

\begin{table}[]
\vspace{-0.4cm}
\centering
\scriptsize
\begin{tabular}{wl{8.5cm}wc{1cm}wc{1cm}}
\hline
\multicolumn{1}{c}{\textbf{Generated instruction}} &
  \multicolumn{1}{c}{\textbf{Score}} &
  \multicolumn{1}{c}{\textbf{Iter. step}} \\ \hline
%\parbox{8.5cm}{\vspace{0.1cm}Identify the main theme or message in the previous text and explain its significance in 2-3 sentences.\vspace{0.1cm}} &
%  4.5 &
%  28 \\\hline
%\parbox{8.5cm}{\vspace{0.1cm}Summarize the previous text in 1 sentence, while also considering the broader context, the author's intent, the potential implications of the information, and also identify the main theme or message and its significance.\vspace{0.1cm}} &
%  4.5 &
%  58 \\\hline
%\parbox{8.5cm}{\vspace{0.1cm}Analyze the previous text and identify the main theme or message, its significance and potential implications, while also considering the broader context, the author's intent, and the impact it may have on the reader, and also provide recommendations for further actions or research.\vspace{0.1cm}} &
%  4.5 &
%  80 \\\hline
\parbox{8.5cm}{\vspace{0.1cm}Summarize the previous text in 2-3 sentences, while also considering the broader context, the author's intent, the potential implications of the information, and also identify the main theme or message and its significance, and also analyze the impact of the information on the reader.\vspace{0.1cm}} &
  5 &
  65 \\\hline
\parbox{8.5cm}{\vspace{0.1cm}Summarize the previous text in 1 sentence, while also considering the broader context, the author's intent, the potential implications of the information, and also identify the main theme or message and its significance, and also analyze the impact of the information on the reader, and also provide recommendations for further\vspace{0.1cm}} &
  5 &
  72 \\\hline
\parbox{8.5cm}{\vspace{0.1cm}Summarize the previous text in 2-3 sentences, while also considering the broader context, the author's intent and the potential implications of the information, and also identify the main theme or message.\vspace{0.1cm}} &
  \textbf{5.5} &
  46 \\ \hline
\end{tabular}
\newline
\caption{\textbf{Top meta-prompting optimized instructions} scoring 5 or higher, with respective scores and iteration steps at which they were obtained, ordered top to bottom, with the top-scoring, retained instruction in the last row.}
\vspace{-1cm}
\label{tab:top_prompts}
\end{table}

It is reasonable to assume that the meta-prompt iteration in subsequent steps used this query and its score in its search for further tentative instructions, with the generated instructions in three subsequent steps (58, 65 and 72) being some derivation of it (second, fourth and fifth rows).

When taking a look at the entire set of generated queries, a high fluctuation of the evaluation scores can be observed along the iteration steps. This is likely due to some interim, generated instructions happening to perform poorly.

\subsection{Examining the responses}

To gain insight about where our method outperformed the plain RAG baseline, we examined the first 10 instances where our method correctly provided the answer while the baseline failed.
Among these, in six cases, the baseline provided a verbose response and might have failed the exact-match evaluation criterion used.\footnote{An example of a verbose response: [query] Was Superhero fiction invented in the digital format? [response] The answer is no; superhero fiction did not originate in digital format. Superheroes have their roots in pulp magazines, comic strips, and comic books, which were all print media formats before the advent of digital technology.} 
% \todo[inline]{Não sei se valerá a pena manter esta footnote longa...R: concordo apaguei.}failing the exact-match evaluation against the correct answer thus resulting in an incorrect response in half of these instances.\todo[inline]{Nesta footnote, este não é um exemplo antes do falhanço do procedimento de avaliação? A resposta correta está lá...Isto não coloca imensamente em dúvida se de facto há superioridade do nosso método?}
In the remaining four cases, the baseline either answered incorrectly, responded with a query, or failed to provide an answer.

Conversely, we reviewed the first 10 instances where our method failed to provide the correct answer while the baseline succeeded. 
We observed that our method exhibited a verbose response behavior in five cases that eventually arrived at the correct answer but failed the exact-match evaluation criterion. 
In two other cases, our method gave a verbose response without providing an answer, while in two remainder cases, it provided an incorrect response. 
Finally, in one instance, our method did not provide any response.

%While these observations do not yield definitive conclusions, empirically, a more accurate response is generated when the retrieved content is transformed with the optimized prompt.
%Although our method answers correctly but verbosely in five cases, 
Both methods seem thus to be similarly penalized by the evaluation criterion for not providing straight answers when the correct answer may happen to be included in the verbose response. % o que quero dizer realmente?

\section{Future work}
\label{sec:future}

%It is important to note that, due to hardware and time constraints, only a limited number of preliminary experiments were conducted, particularly with the meta-prompt definition.

While providing a method that effectively enhances RAG, our proposal paves the way for future research, such as the exploration of optimal hyper-parameters, refining content retrieved without gold content, scaling up with larger models, exploring further evaluation functions, or tackling other downstream tasks.
A significant number of hyper-parameters remain unexplored, which is an opportunity to further enhance the efficacy of this method.

It is worth noting that the evaluation with exact matching is a binary task, and achieving an exact match with a task demanding a more complex string match still needs to be studied, questioning the need for additional training, a different meta-prompt, or a different approach.

% It will be interesting also to study the interaction of our proposed method, on optimizing the retrieval instruction, and the optimization of the query itself. Exploratory experiments seem to indicate that each of these may get at some correct answers that the other one does not. Finding ways to leverage each other's strengths will be important.

It will be interesting also to study the interaction of our proposed method with the Portuguese language \cite{osorio-etal-2024-portulan,branco-etal-2020-infrastructure} with the existing family of LLMs \cite{rodrigues2023advancing,santos-etal-2024-advancing,santos-etal-2024-fostering} and multi-modal LLMs \cite{santos-etal-2022-cost} as also with other tasks such as argument mining \cite{rodrigues-branco-2022-transferring}, exploring data spuriousness and others \cite{branco-etal-2021-shortcutted,santos2021neural,branco2020comparative,rodrigues2018predicting}.

\section{Conclusion}
\label{sec:conclusion}

This paper introduces a novel method that enhances RAG and that can be combined with previous approaches for RAG enhancement. It consists in refining the retrieved content included in the prompt entered into the generation model with the help of a refinement instruction that was obtained by means of meta-prompting optimization.

It reports also on the empirical assessment of this proposal by means of it being embedded in a most demanding multi-hop question answering task. The evaluation results indicate that it is highly effective in as much as it outperforms RAG without this method by over 30\%.

%While human-designed prompts may still outperform our method, it is important to recognize the advantages of automation and adaptability to new data found in our method.

\begin{credits}
\subsubsection{\ackname} 
This research was partially supported by:
PORTULAN CLARIN — Research Infrastructure for the Science and Technology of Language, funded by Lisboa 2020, Alentejo 2020 and FCT (PINFRA/22117/2016); ACCELERAT.AI - Multilingual Intelligent Contact Centers, funded by IAPMEI (C625734525-00462629);
\end{credits}

\bibliographystyle{splncs04}
\bibliography{bibliography}
\end{document}